\documentclass[10pt,twocolumn,letterpaper]{article}

\usepackage{wacv}
\usepackage{times}
\usepackage{epsfig}
\usepackage{graphicx}
\usepackage{amsmath}
\usepackage{amssymb}
\usepackage{booktabs}
\usepackage{amsfonts,amsthm}
\usepackage{multirow}
\usepackage{float}
\usepackage{caption}
\usepackage{enumitem}
\usepackage{color}

\makeatletter
\newcommand{\oset}[3][0ex]{%
  \mathrel{\mathop{#3}\limits^{
    \vbox to#1{\kern-1\ex@
    \hbox{$\scriptstyle#2$}\vss}}}}
\makeatother

\makeatletter
\newcommand{\ooset}[3][0ex]{%
  \mathrel{\mathop{#3}\limits^{
    \vbox to#1{\kern0.05\ex@
    \hbox{$\scriptstyle#2$}\vss}}}}
\makeatother

\def\wacvPaperID{729} 

\wacvfinalcopy 

\ifwacvfinal
\usepackage[breaklinks=true,bookmarks=false]{hyperref}
\else
\usepackage[pagebackref=true,breaklinks=true,colorlinks,bookmarks=false]{hyperref}
\fi

\begin{document}

\title{Cross-identity Video Motion Retargeting with \\Joint Transformation and Synthesis}

\author{
Haomiao Ni$^1$\qquad\qquad Yihao Liu$^2$\qquad\qquad Sharon X. Huang$^1$\qquad\qquad Yuan Xue$^2$ \\
$^1$The Pennsylvania State University, University Park, PA, USA \\ 
$^2$Johns Hopkins University, Baltimore, MD, USA \\
$^1${\tt\small\{hfn5052, suh972\}@psu.edu}\quad$^2${\tt\small\{yliu236, yuanxue\}@jhu.edu}
}

\twocolumn[{%
\renewcommand\twocolumn[1][]{#1}%
\maketitle
\begin{center}
\vspace{-5pt}
    \centering
    \includegraphics[width=0.95\linewidth]{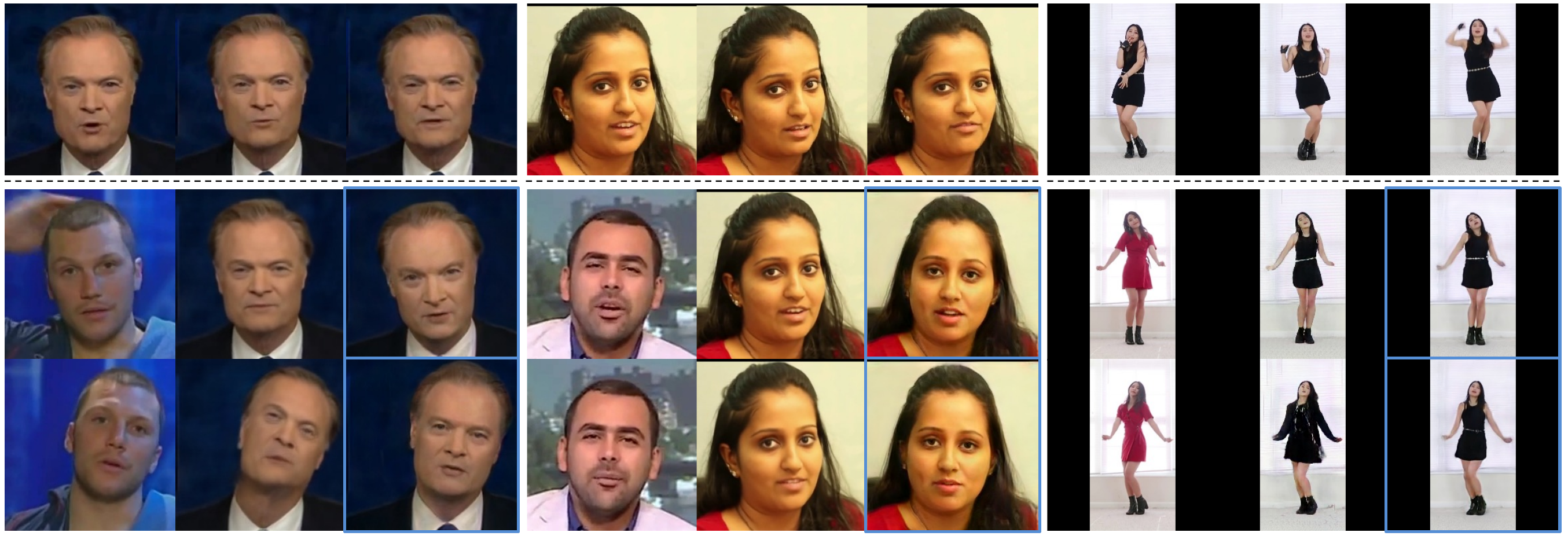}
    \captionof{figure}{Examples of video motion retargeting, where motion from the driving video (1st column in 2nd\&3rd rows) is transferred to the subject in the subject video (1st row).  Videos generated by RegionMM \cite{siarohin2021motion} for face video and EDN \cite{chan2019everybody} for dance video are shown in the 2nd column of each block, 2nd\&3rd rows.  Videos generated by our proposed TS-Net are in the 3rd column of each block, 2nd\&3rd rows (highlighted with {\color{blue}blue} boxes). 
    }
    \label{fig:examples}
\end{center}%
}]

\begin{abstract}
\vspace{-9.5pt}
In this paper, we propose a novel dual-branch Transformation-Synthesis network (TS-Net), for video motion retargeting.
Given one subject video and one driving video, TS-Net can produce a new plausible video with the subject appearance of the subject video and motion pattern of the driving video. TS-Net consists of a warp-based transformation branch and a warp-free synthesis branch.
The novel design of dual branches combines the strengths of deformation-grid-based transformation and warp-free generation for better identity preservation and robustness to occlusion in the synthesized videos. A mask-aware similarity module is further introduced to the transformation branch to reduce computational overhead. Experimental results on face and dance datasets show that TS-Net achieves better performance in video motion retargeting than several state-of-the-art models as well as its single-branch variants. Our code is available at \url{https://github.com/nihaomiao/WACV23_TSNet}.
\end{abstract}

\section{Introduction}
\vspace{-5pt}
Motion retargeting aims to transfer motion from a driving video to a target video while maintaining the subject's identity of the target video.
It has become an important topic due to its practical applications in special effects, virtual/augmented reality and video editing, \etc. Motion retargeting in the image domain has been explored extensively and compelling results have been shown in many tasks, such as person image generation \cite{balakrishnan2018synthesizing, ma2017pose, pumarola2018unsupervised,siarohin2019appearance, siarohin2021motion}, and facial expression generation \cite{chen2020puppeteergan, kim2019u, ren2021pirenderer,zhang2020cross}.
Often formulated as a guided video synthesis task, motion retargeting between videos is known to be more challenging than motion retargeting between images since the temporal dynamics of the motion to be transferred has to be learned~\cite{chu2020learning}. Moreover, synthesizing realistic videos, especially human motion videos, is more challenging than the generation of high-quality images because human perception is sensitive to unnatural temporal changes, and human motion is often highly articulated~\cite{yang2020transmomo,zakharov2019few}. In this paper, we mainly focus on video motion retargeting between different human subjects (Fig.~\ref{fig:examples}). Given one subject video and one driving video, we aim to synthesize a new plausible video with the same identity of the person from the subject video and the same motion as the person in the driving video.  

Recent works in video motion retargeting \cite{bansal2018recycle,chan2019everybody,chu2020learning,gafni2021dynamic, hong2022depth, jeon2020cross, wang2019few,wang2018video, wang2021one, wiles2018x2face,yang2020transmomo,zakharov2019few} have shown impressive progress.
To capture the temporal relationship among video frames, prior works \cite{chu2020learning,wang2021one, wiles2018x2face} generated frames via warping subject frames by motion flow, which is usually extracted by specifically designed warping field estimators, such as FlowNet \cite{dosovitskiy2015flownet} or first-order approximation \cite{siarohin2020first}. While warp-based systems can generally preserve subject identity well, traditional flow-based warping may suffer from occlusion and large motion due to its requirement of learning a warp field with point-to-point correspondence between frames \cite{hu2018videomatch}. Other methods \cite{bansal2018recycle,chan2019everybody,ha2020marionette,kim2018deep, yang2020transmomo,zakharov2019few} utilized warp-free (direct) synthesis with a conditional GAN-style structure \cite{isola2017image,mirza2014conditional,wang2018high}. To ease the challenging of direct synthesis, they often employed feature disentangle/decomposition \cite{yang2020transmomo} or followed the state-of-the-art generator architectures \cite{park2019semantic,wang2018high} to add various connections among inputs, the encoder, and the decoder network.
Unlike warp-based generation, direct synthesis is not limited to only using pixels from reference images, and therefore is easier to synthesize novel pixels for unseen/occluded objects. However, such flexibility can also lead to identity leakage  \cite{ha2020marionette}, \textit{i.e.}, identity changes in the generated video. 

Considering that warp-based synthesis can better preserve identity while warp-free generation helps produce new pixels, in this paper, we propose a novel video motion retargeting framework, termed \textit{Transformation-Synthesis Network}, or \textit{TS-Net} for short, to combine their advantages. TS-Net has a dual branch structure which consists of a transformation branch and a synthesis branch. 
The network architectures within the two branches are inherently different, thus learning via the two branches can be regarded as a special multi-view learning case \cite{xu2013survey}.
Unlike the popular warp-based methods using specially designed optical flow estimators \cite{siarohin2019animating,siarohin2020first, siarohin2021motion} and inspired by~\cite{liu2022coordinate},
our proposed transformation branch computes deformation flow by weighting the regular grid with a spatial similarity matrix between driving mask features and subject image features. The computation of similarity takes multiple correspondences into consideration; thus it can better alleviate occlusion and handle large motion. 
We also design a mask-aware similarity to avoid comparing all pairs of points within the feature maps and thus be more efficient than traditional similarity computation methods.
In our synthesis branch, we use a fully-convolutional fusion network.  Features of two branches are concatenated and fed to the decoder network to generate realistic video frames. Experiments in Sec. \ref{sec:exp} shows the effectiveness of this simple concatenation strategy.
 
Merely based on sparse 2D masks of driving videos, our proposed TS-Net can consistently achieve state-of-the-art results for both face and dance videos, successfully modeling hair and clothes details and their motion. TS-Net also handles large motions and preserves identity better when compared with other state-of-the-art methods, as shown in Fig.~\ref{fig:examples}. 
Our contributions are summarized as follows:
\begin{enumerate}[topsep=0pt, partopsep=0pt, itemsep=1pt, parsep=1pt]
    \item We propose a novel dual branch video motion retargeting network TS-Net to generate identity-preserving and temporally coherent videos via joint learning of transformation and synthesis. 
    
    \item We utilize a simple yet effective way to estimate deformation grid based on similarity matrix. Mask-aware similarity is adopted to further reduce computation overhead. 
    
    \item Comprehensive experiments on facial motion and body motion retargeting tasks show that TS-Net can achieve state-of-the-art results by only using sparse 2D masks. 
\end{enumerate}

\begin{figure*}[t]
    \centering
    \includegraphics[width=0.90\linewidth]{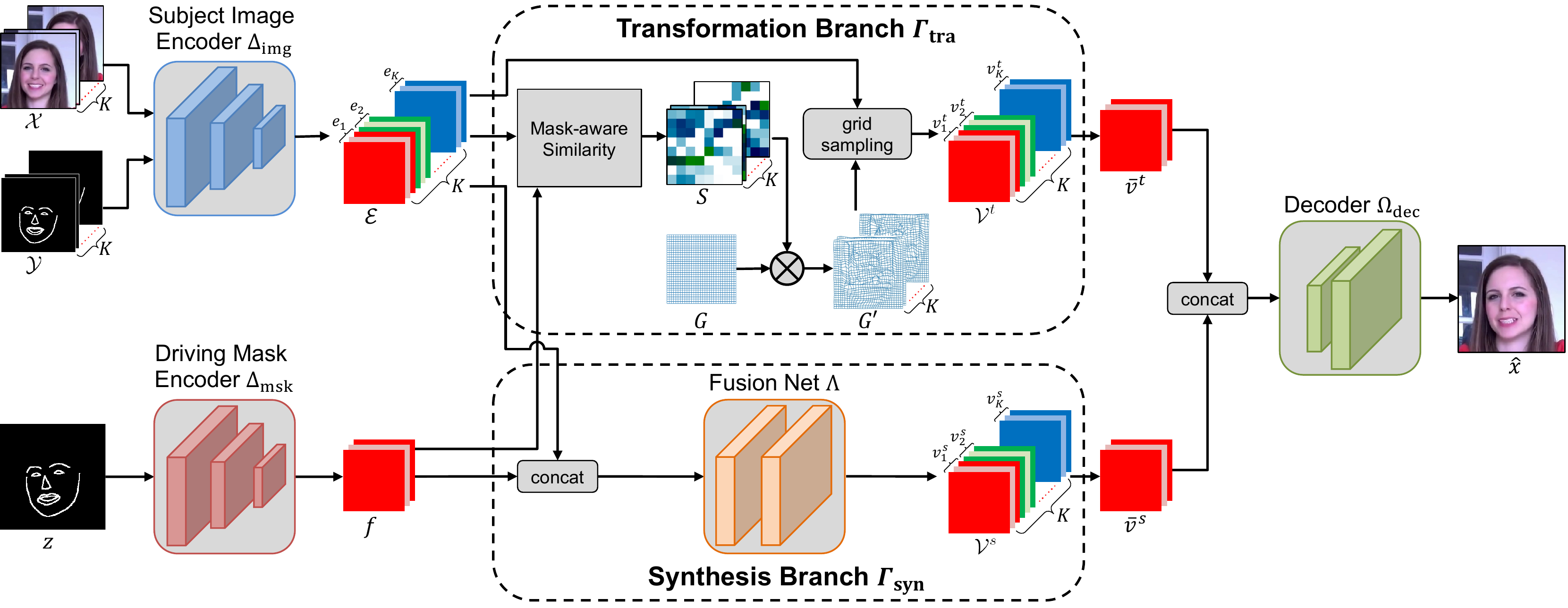}
    \caption{Illustration of the TS-Net generator to generate one frame $\hat{x}$ in the target video.}
    \label{fig:arch}
\end{figure*}

\section{Related Work}
\noindent\textbf{Guided Image Generation.} For conditional image generation, many works focused on generation tasks guided by specific conditions such as pose-guided person image synthesis \cite{balakrishnan2018synthesizing,ma2017pose,pumarola2018unsupervised,sarkar2020neural,siarohin2018deformable,song2019unsupervised} and conditioned facial expression generation \cite{chen2020puppeteergan,ha2020marionette,ren2021pirenderer}.
Pose-guided person image generation can produce person images in arbitrary poses, based on a subject image of that person and a novel pose from the driving image. 
Ma \etal \cite{ma2017pose} proposed a two-staged coarse-to-fine Pose Guided Person Generation Network ($\rm{PG}^2$), which utilizes pose integration and image refinement to generate high-quality person images.
Conditioned facial expression generation aims to generate a reenacted face which shows the same expression as the driving face image while preserving the identity of the subject image.
Chen \etal \cite{chen2020puppeteergan} proposed a two-stage framework called PuppeteerGAN, which first performs expression retargeting by the sketching network and then executes appearance transformation by the coloring network.
Though these works have shown promising results, they are restricted to a specific object category (face or human body). Several recent works \cite{siarohin2019appearance,siarohin2019animating,siarohin2021motion,zhang2020cross} have proposed general guided image generation in various domains. 
Most of works \cite{siarohin2019animating,siarohin2020first,siarohin2021motion,tao2022structure} applies motion flow to image animation because it can model the physical dynamics.
Siarohin \etal \cite{siarohin2020first} proposed a general self-supervised first-order-motion model for estimating dense motion flow to animate arbitrary objects using learned keypoints and local affine transformations. In \cite{siarohin2021motion}, the authors further improved their network by modeling object movement through unsupervised region detection. 
Despite of building upon similar motion flow, instead of adopting complicated modeling in \cite{siarohin2020first,siarohin2021motion}, the transformation branch in TS-Net generate deformation flow by weighting regular grid with similarity matrix in feature space, which shows better simplicity and efficiency.

\noindent\textbf{Video Motion Retargeting.} 
Different from image-based generation, video motion retargeting is more challenging due to the additional coherence requirements in the temporal dimension.
Most existing literature focused on specific domains such as human pose motion retargeting \cite{chan2019everybody, yang2020transmomo}, or facial expression retargeting \cite{gafni2021dynamic,ha2020marionette,kim2018deep,wang2021one,wiles2018x2face,zakharov2019few}, yet they may lack generality when applied to multiple domains.
In contrast, our proposed TS-Net can work well on both face and human body videos.
Using off-the-shelf detectors to extract driving motion masks, such as 3D masks \cite{gafni2021dynamic,ha2020marionette,kim2018deep}, 2D dense mask \cite{wang2019few,wang2018video}, or 2D sparse mask \cite{chan2019everybody,zakharov2019few}, is also popular in current video motion retargeting methods. Due to the simplicity of 2D sparse masks, our proposed TS-Net also utilizes keypoints extracted by Dlib \cite{king2009dlib} and OpenPose \cite{cao2019openpose} to synthesize videos of 3D human face/body.
To learn representation and preserve input information effectively, 
most recent methods are based on state-of-the-art generators with U-Net structure and AdaIN module \cite{ha2020marionette,kim2018deep,wang2019few,zakharov2019few}, feature disentangle/decomposition \cite{wang2021one,yang2020transmomo}, or specifically designed motion flow estimators \cite{chu2020learning,wang2019few,wiles2018x2face}. On the contrary, our proposed TS-Net uses a more robust and general GAN generator \cite{johnson2016perceptual} as backbone to jointly learn transformation and synthesis.
Some previous works \cite{wang2019few,wang2018video} also performed video motion retargeting by combining warp-based and warp-free generation. However, their warping flows are always applied to previous generated frames, which may lead to the accumulation of synthesis artifacts. Our proposed TS-Net instead computes the warping flow between driving mask and \textit{real} subject images in feature space to avoid this issue.

\section{Methodology}
\subsection{Model Architecture}
Given a sequence $\mathcal{X} = \{x_1, x_2, \dots, x_K\}$ with $K$ subject frames, their corresponding mask sequence $\mathcal{Y}=\{y_1, y_2, \dots, y_K\}$, and a mask frame $z$ from driving video, the TS-Net generator can produce a new video frame $\hat{x}$ with the subject from $\mathcal{X}$ and mask from $z$.
Masks are generated by applying off-the-shelf pretrained 2D sparse keypoint detectors, \textit{i.e.}, Dlib \cite{king2009dlib} for face landmark detection and OpenPose \cite{cao2019openpose} for pose keypoint estimation.
As illustrated in Fig.~\ref{fig:arch}, TS-Net generator consists of two branches: a transformation branch $\Gamma_\text{tra}$ and a synthesis branch $\Gamma_\text{syn}$ for generating the new video frame using warp-based transformation and direct synthesis, respectively. 

During training, we concatenate $K$ subject frames $\mathcal{X}$ and their masks $\mathcal{Y}$ and feed them to an image encoder $\Delta_\text{img}$ to extract subject embedding features $\mathcal{E}=\{e_1, e_2, \dots, e_K\}$. 
A mask encoder $\Delta_\text{msk}$ encodes the input driving mask $z$ into driving embedding feature $f$. 
To reduce computational costs of matrix multiplication, TS-Net operates in a low-resolution feature space, where the spatial size of $\mathcal{E}$ and $f$ are only 1/$8^2$ of the input frames. We then input $\mathcal{E}$ and $f$ to the transformation branch $\Gamma_\text{tra}$ and the synthesis branch $\Gamma_\text{syn}$, as illustrated as follows.

\noindent\textbf{Transformation Branch.}
Inside $\Gamma_\text{tra}$, we implement warp-based transformation using spatial sampling grids \cite{jaderberg2015spatial}. We first compute the cosine similarity matrix $S_k$ between the driving embedding feature $f$ and the $k$-th subject feature $e_k$ as
\begin{small}
\begin{equation}
\label{eq:cos}
    S_{k_{pq}}=  \frac{e_{k_p}\cdot f_{q} }{\left\|e_{k_p}\right\|_2\left\|f_{q}\right\|_2}\enspace,
\end{equation}
\end{small}
\vspace{-2mm}

\noindent where $S_{k_{pq}}$ is the affinity value between $f_{q}$ at position $q$ in map $f$, and, $e_{k_p}$ at position $p$ in map $e_k$, and $\left\|\cdot\right\|_2$ indicates the L2 norm. 
{Suppose that the size of feature $f$ and $e_k$ are $m\times m$,
the size of matrix $S_k$ will be $m^2\times m^2$, which is quartic to $m$. Thus adopting low-resolution feature maps is important to alleviate computational overhead.} 

\begin{figure}[t]
    \centering
    \includegraphics[width=0.95\linewidth]{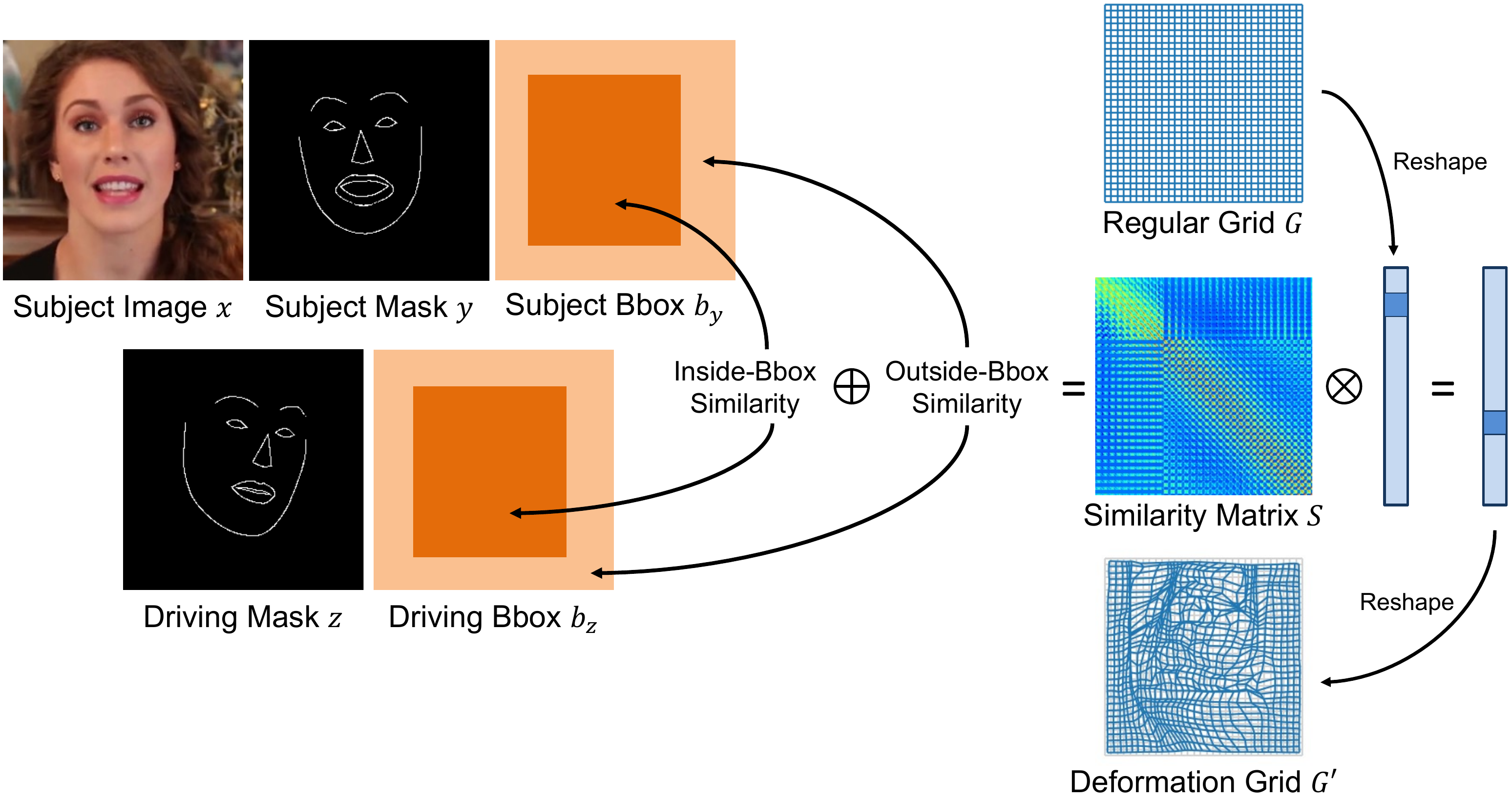}
    \caption{Illustration of our proposed mask-aware similarity computation.}
    \label{fig:sim}
\end{figure}

We further reduce computational costs by designing a novel mask-aware similarity computation method, as shown in Fig.~\ref{fig:sim}. Given one driving mask $z$ and one subject mask $y$, we first generate their corresponding bounding box $b_z$ and $b_y$ according to the maximum and minimum keypoint coordinates in masks. Intuitively, most of pixels inside the bounding box $b_z$ will not be warped to the regions outside $b_y$ thus we can skip similarity computation between pixels of these two regions.
Based on this observation, we downsample $b_z$ and $b_y$ to be the same spatial size as feature map $f$ and $e$ and then only compute the affinity values between points of their inside/outside-bounding-box regions.

For the input subject features $\mathcal{E}$, we now have $K$ similarity matrices $S=\{S_1, S_2, \dots, S_K\}$.
We then use similarity matrix $S_k$ to weight the regular grid $G$ and obtain the $k$-th sampling grid $G'_k$ as
\begin{small}
\begin{equation}
\label{eq:warp}
    G'_{k_p} = \frac{\sum\nolimits_{q}\left(\exp{(\tau S_{k_{pq}})}\cdot G_{q}\right)}{\sum\nolimits_{q}\exp{(\tau S_{k_{pq}})}}\enspace,
\end{equation}
\end{small}
\vspace{-2mm}

\noindent where $G'_{k_p}$ is the coordinate $p$ of sampling grid $G'_{k}$, $G_{q}$ is the coordinate $q$ of regular grid $G$, and {$\tau$ is the coefficient to control the relative difference between affinity values.} This results in $K$ sampling grids ${G}'=\{G'_1, G'_2, \dots, G'_K\}$. By applying sampling grids $G'$ to subject features $\mathcal{E}$, 
we acquire $K$ warped features $\mathcal{V}^t=\{v^t_1, v^t_2, \dots, v^t_K\}$.
The final warped feature $\bar{v}^t$ is then generated by averaging the $K$ features in $\mathcal{V}^t$.

\noindent\textbf{Synthesis Branch.}
Inside $\Gamma_\text{syn}$, we concatenate the $k$-th subject embedding feature $e_k$ with driving mask feature $f$ and feed them to a fusion network $\Lambda$, which consists of a series of fully-convolutional layers, for creating the $k$-th synthesized warp-free feature map $v^s_k$. Processing $K$ feature maps in $\mathcal{E}$ will generate $K$ synthesized feature maps $\mathcal{V}^s=\{v^s_1, v^s_2, \dots, v^s_K\}$. We then take average of the $K$ features in $\mathcal{V}^s$ to produce the final synthesized feature $\bar{v}^s$.

\noindent\textbf{Combination of Branches.}
We concatenate the feature $\bar{v}^t$ and $\bar{v}^s$ of two branches and adopt a decoder network $\Omega_\text{dec}$ to synthesize the final output $\hat{x}$. 
We also tried to combine $\bar{v}^t$ and $\bar{v}^s$ with an attention-based matting function as in \cite{wang2019few,wang2018video}, yet we found that this strategy fails to generate better results, as later illustrated in Sec.~\ref{sub_sec:exp_res}.
More architecture details are in Sec.~\ref{sec:implementation}.

\subsection{{Training and Inference}}
We train our proposed TS-Net generator using a self-supervised way of training. Specifically, input driving mask sequence $\mathcal{Z}$ and subject image sequence $\mathcal{X}$ are from different segments of the same subject video. Thus we have frames in the subject video as ground truth. The overall loss for generating one frame is calculated as
\begin{small}
\begin{equation}
\label{eq:same}
    l = \mathcal{L}_\text{GAN}(\hat{x}, x) + \alpha \mathcal{L}_\text{VGG}(\hat{x}, x) + \beta \mathcal{L}_\text{FM}(\hat{x}, x) + \lambda \mathcal{L}_\text{TRA}(\hat{x}^t, x)\enspace,
\end{equation}
\end{small}

\noindent where $\mathcal{L}_\text{GAN}$ is an adversarial loss \cite{goodfellow2014generative}, $\mathcal{L}_\text{VGG}$ represents a perceptual loss \cite{johnson2016perceptual} based on VGG network \cite{simonyan2014very}, $\mathcal{L}_\text{FM}$ is a feature matching loss \cite{wang2018high}, and $\mathcal{L}_\text{TRA}$ is extra regularization loss for transformation branch. Here $\alpha$, $\beta$, and $\lambda$ are balancing factors, $\hat{x}$ is the generated frame, $\hat{x}^t$ is warped subject frame, and $x$ is ground truth real frame. 

We now introduce detailed loss terms. The adversarial loss $\mathcal{L}_\text{GAN}$ is defined by the minimax optimization \cite{goodfellow2014generative}:
\begin{small}
\begin{equation}
\label{eq:adv}
    \min_{G}\max_{D}~\mathbb{E}_{x}[\log D(x)] + \mathbb{E}_{\hat{x}}[\log(1-D(\hat{x}))]\enspace.
\end{equation}
\end{small}
\vspace{-2mm}

\noindent Discriminator $D$ is designed to distinguish the real video frame $x$ from the synthesized video frame $\hat{x}$ given driving mask frame $z$. The perceptual loss $\mathcal{L}_\text{VGG}$ is defined as
\begin{small}
\begin{equation}
    \sum^{N}_{i=1}\frac{1}{W_i}\left[||F^{(i)}(\hat{x})-F^{(i)}(x)||_{1}\right]\enspace,
\end{equation}
\end{small}
\vspace{-2mm}

\noindent where $N$ is the number of layers in VGG feature extraction network and $F^{(i)}$ denotes the output of $i$-th layer with $W_{i}$ elements of the VGG network~\cite{simonyan2014very} pretrained on ImageNet~\cite{deng2009imagenet}.
The feature matching loss $\mathcal{L}_\text{FM}$ is defined as
\begin{small}
\begin{equation}
    \sum^{M}_{i=1}\frac{1}{U_i}\left[||D^{(i)}(\hat{x})-D^{(i)}(x)||_{1}\right]\enspace,
\end{equation}
\end{small}
\vspace{-2mm}

\noindent where $D^{(i)}$ denotes the $i$-th layer with $U_i$ elements of our proposed discriminator $D$. 
The transformation branch loss $\mathcal{L}_\text{TRA}$ is calculated as
\begin{small}
\begin{equation}
    \mathcal{L}_\text{TRA} = ||\hat{x}^t-x||_{1}\enspace,
\end{equation}
\end{small}
\vspace{-2mm}

\noindent where $\hat{x}^t$ is computed by patch-wise warping subject frame using deformation grid $G'$. For $K$ input subject frames, we compute $\mathcal{L}_\text{TRA}$ for each frame and then sum them up. In \eqref{eq:same}, the first three loss terms ($\mathcal{L}_\text{GAN}$, $\mathcal{L}_\text{VGG}$, and $\mathcal{L}_\text{FM}$) are commonly used in current video generation models \cite{wang2018high,wang2018video}.  We show the importance of introducing $\mathcal{L}_\text{TRA}$ to the training of our model in Sec.~\ref{sub_sec:exp_res}.

\begin{figure*}[ht]
    \centering
    \includegraphics[width=0.9\textwidth]{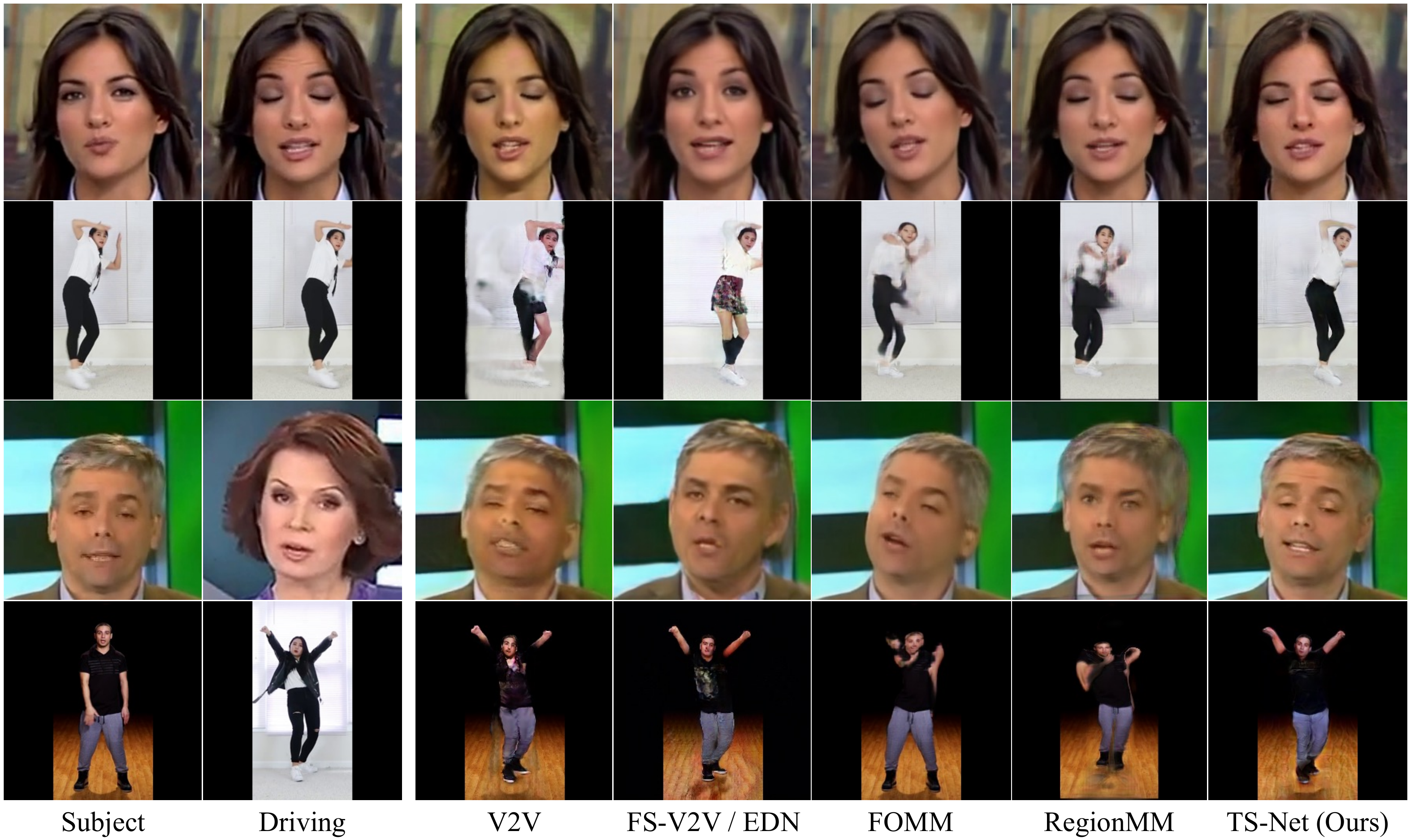}
    \caption{Qualitative comparison with state-of-the-art methods (V2V \cite{wang2018video}, FS-V2V \cite{wang2019few}, EDN \cite{chan2019everybody}, FOMM \cite{siarohin2020first},
    and RegionMM \cite{siarohin2021motion}) on face and dance video datasets. The top two rows are results for self-reconstruction and the bottom two rows are for cross-identity transfer. Note that FS-V2V is used for face videos and EDN is used for dance videos.}
    \label{fig:sota}
\end{figure*}

\noindent\textbf{{Inference.}}
Given the subject video $\mathcal{X}$ and the mask sequence of driving video $\mathcal{Z}$, we randomly select $K$ frames from the subject video to synthesize a new frame $\hat{x}$. 

\section{Experiments}
\label{sec:exp}
\subsection{Datasets and Metrics}
\noindent\textbf{Datasets.} We conduct experiments on face videos and dance videos. For face videos,
    We use the real videos in FaceForensics \cite{roessler2018faceforensics} dataset, which contains 1,004 videos of news briefing from different reporters. 
    We randomly choose 150 videos for training 
    and 150 videos for testing. Since the original videos are long, we randomly selected a short segment of 30 continuous frames from each video, and the selected short videos are used in our experiments. 
    To extract mask sequences from videos, we first apply a face alignment algorithm \cite{king2009dlib} to localize 68 facial landmarks in each frame. The sparse facial landmarks are then connected to create the face mask.
    For dance videos, following \cite{ wang2019few,wang2018video}, we downloaded dancing videos from Youtube\footnote{The video links are available on the \href{https://github.com/NVlabs/few-shot-vid2vid/blob/master/data/preprocess/youTube_playlist.txt}{project website} of \cite{wang2019few}. We obtained permission to use the videos from the video owners.}. We randomly chose 100 videos for training and 85 videos for testing and randomly sampled 30 continuous frames containing only one person from each video. We extracted human poses as masks via OpenPose \cite{cao2019openpose}. Face and hand keypoints are kept for better motion retargeting. 

\noindent
{\textbf{Metrics.}} Following \cite{siarohin2019animating,siarohin2020first},
we compute metrics based on two testing settings, \textit{self-reconstruction} and \textit{cross-identity transfer}. For each setting, we synthesize 100 videos where the size of each frame is $256\times 256$ .
    For \textit{self-reconstruction}, we segment a video of the same subject to two non-overlapping clips and use one clip as subject video and another one as driving video. In this setting, driving video also serves as ground truth. Similar to \cite{gafni2021dynamic}, we compute the normalized mean $L_2$ distance and Learned Perceptual Image Patch Similarity (LPIPS) \cite{zhang2018perceptual} metrics between self-reconstructed results and driving videos.
    For \textit{cross-identity transfer}, which is more practical in real world applications, subject video and driving video are from different subjects in this setting. Due to the lack of ground truth, we conduct user study to compare our models with state-of-the-art methods. Human evaluators are shown sets of $n$ videos generated by $n$ different models and then are asked to rank videos in each set from 1 (best) to $n$ (worst) based on perceptual similarity and realism. Tied rank scores will be given for videos that are perceived to have comparable quality. 

\begin{table}[t]
\centering
\resizebox{0.7\linewidth}{!}{%
\begin{tabular}{c|l|c|c}
\hline
Dataset                & Method        & $L_2\downarrow$     & LPIPS $\downarrow$  \\
\hline
\multirow{7}{*}{Face}  & V2V \cite{wang2018video}           & 0.0356 & 0.1123 \\
                      & FS-V2V \cite{wang2019few}       & 0.0422 & 0.1064 \\
                      & FOMM \cite{siarohin2020first}        & 0.0443 & 0.1184 \\
                      & RegionMM \cite{siarohin2021motion}
                      & \textbf{0.0148} & \textbf{0.0532} \\
                      \cline{2-4} 
                      & TS-Net ($K=1$) & 0.0275 & 0.0731 \\
                      & TS-Net ($K=3$) & 0.0271 & 0.0683 \\
                      & TS-Net ($K=5$) & {0.0270} & {0.0673}\\
\hline
\multirow{7}{*}{Dance} & V2V \cite{wang2018video}          & 0.0895 & 0.2622 \\
                      & EDN \cite{chan2019everybody}          & 0.0471 & 0.1718 \\
                      & FOMM \cite{siarohin2020first}         & 0.1517 & 0.3081 \\
                      & RegionMM \cite{siarohin2021motion}
                      & 0.1945 & 0.4081 \\
                      \cline{2-4} 
                      & TS-Net ($K=1$) & 0.0433 & 0.1586 \\
                      & TS-Net ($K=3$) & \textbf{0.0421} & 0.1543 \\
                      & TS-Net ($K=5$) & {0.0423} & \textbf{0.1541} \\
\hline
\end{tabular}%
}
\caption{Comparison with state-of-the-art methods under the self-reconstruction setting on face and dance datasets. $K$ is the number of subject frames used in generation.}
\label{tab:sota}
\vspace{-5pt}
\end{table}

\subsection{Implementation}\label{sec:implementation}
\noindent
\textbf{Model Implementation.}
Our proposed encoder $\Delta$ and decoder $\Theta$ in TS-Net are general and can have various backbone networks, such as pix2pix~\cite{isola2017image} and SPADE~\cite{park2019semantic}. We adopt the architecture in \cite{johnson2016perceptual} due to its simplicity. For the encoder $\Delta_{\text{img}}$, we use the network with three stride-2 convolutions and 9 residual blocks \cite{he2016deep}. For $\Delta_{\text{msk}}$, we use three stride-2 convolutions without additional residual blocks since masks contain less information. Thus the spatial size of embedding feature map is only 1/$8^2$ size of input image.
To encode position-related information for better synthesis, we apply coordinate convolution \cite{liu2018intriguing} to inputs. For decoder $\Theta_{\text{dec}}$, we employ 4 residual blocks, followed by three up-sampling and convolution layers. For fusion network $\Lambda$, we use one residual block and one $1\times 1$ convolution \cite{lin2013network} to generate warp-free feature maps $\mathcal{V}^s$.
Instance normalization \cite{ulyanov2016instance} is adopted in TS-Net. 
For the discriminator $D$, we use $70\times 70$ PatchGAN \cite{isola2017image,wang2018high,zhu2017unpaired}, which aims to classify whether the $70\times 70$ overlapping patches are real or fake. To stabilize the training, we use LSGAN \cite{mao2017least} for the adversarial loss. 

When training TS-Net, we set batch size as 20 videos and train the model for 600 epochs using the Adam optimizer \cite{kingma2014adam} with $(\beta_1, \beta_2)=(0.5, 0.999)$. The learning rate is fixed to $2\times10^{-4}$ in the first 275 epochs and then linearly decayed to zero. The balancing parameters $\alpha$, $\beta$, and $\lambda$ are all set to be 10 in \eqref{eq:same}. The coefficient $\tau$ in \eqref{eq:warp} is set to be 100.
Data augmentation such as color jitter and flipping are also applied. 
Hyper-parameters are selected via multiple runs of experiments. 
When training our models on the face video dataset, we adopt an image gradient difference loss \cite{mathieu2015deep} as an extra smoothness constraint to eliminate minor artifacts in the generated videos. When training our models on the dance video dataset, similar to \cite{chan2019everybody, wang2019few, wang2018video}, we introduce an extra face discriminator to synthesize better face details. To normalize masks across different subjects,
the masks of driving videos are aligned to the masks of subject videos with the similar methods used in \cite{chan2019everybody,wang2019few}. 

\noindent
\textbf{Baselines.} 
For face video dataset, we choose four state-of-the-art video synthesis or image animation models, vid2vid (V2V) \cite{wang2018video}, few-shot vid2vid (FS-V2V) \cite{wang2019few}, FOMM \cite{siarohin2020first}, and RegionMM \cite{siarohin2021motion} as baselines. For dance video dataset, we compare TS-Net with V2V, FOMM, RegionMM, and Everybody Dance Now (EDN) \cite{chan2019everybody}.
FS-V2V is not included for dance videos since it requires DensePose \cite{guler2018densepose} as extra inputs. We follow the default settings in the methods' original implementations wherever possible 
 . The original V2V and EDN train with a single video and test on the same video. For fair comparison, we train V2V and EDN using all available training videos. 

\begin{figure}[t]
    \centering
    \includegraphics[width=\linewidth]{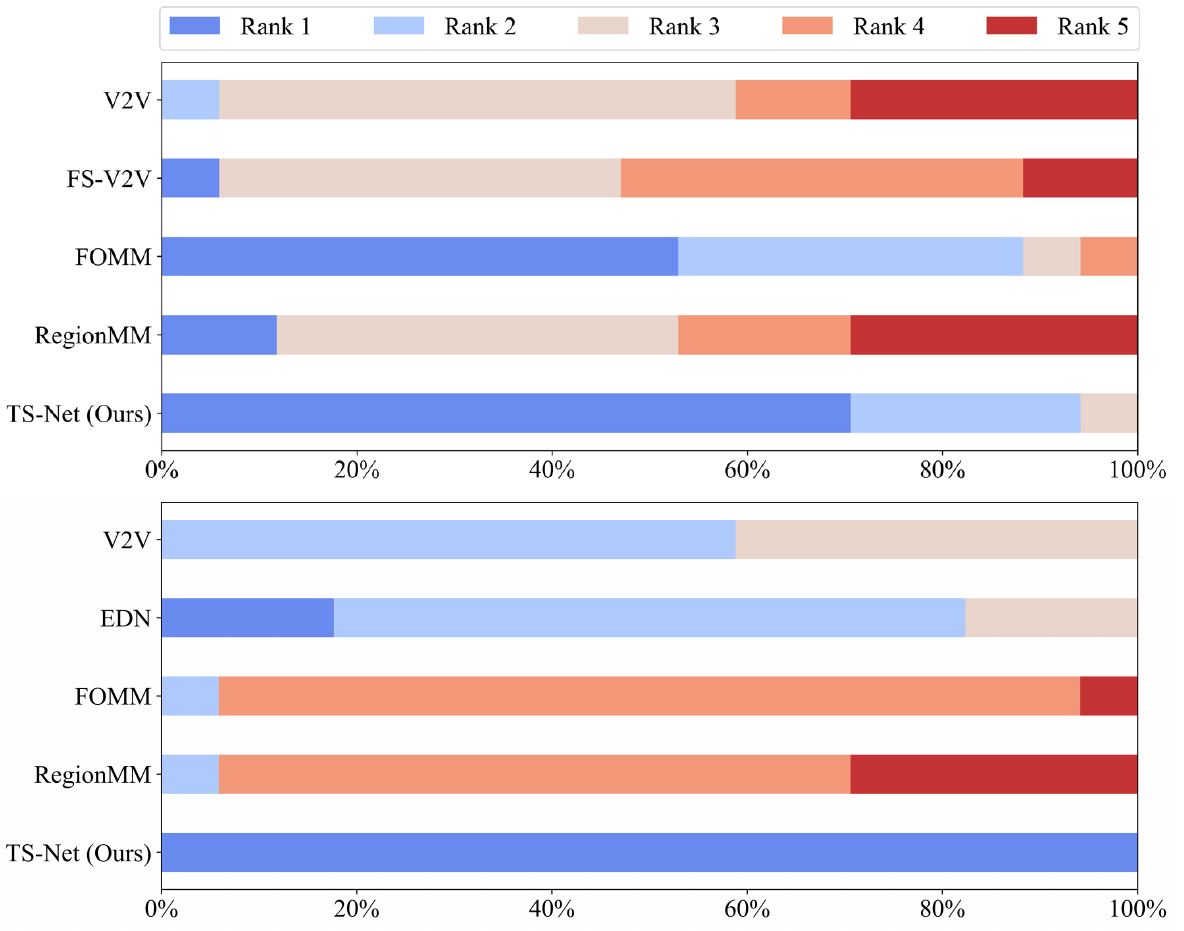}
    \caption{User study of ranking different methods under the cross-identity setting. Ties are allowed. The top chart is for face videos and the bottom one is for dance videos.}
    \label{fig:user}
\vspace{-2pt}
\end{figure}
\subsection{{Result Analysis}}
\label{sub_sec:exp_res}

\begin{figure}[t]
    \centering
    \includegraphics[width=\linewidth]{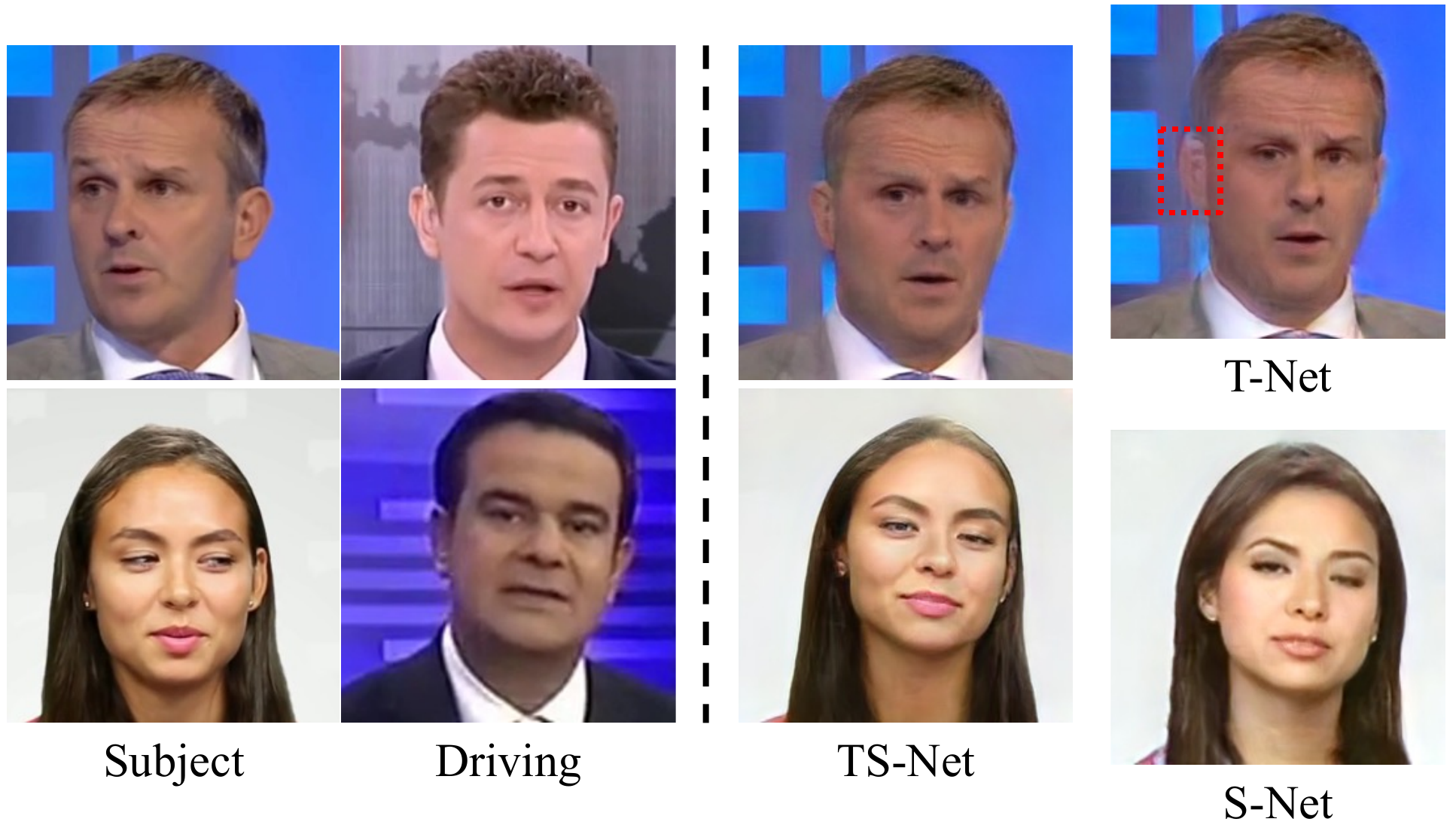}
    \caption{Ablation study under the cross-identity transfer setting on the face dataset.}
    \label{fig:aba-face}
\vspace{-3pt}
\end{figure}

\noindent
\textbf{Comparison with state-of-the-art methods.} Table~\ref{tab:sota} shows a quantitative comparison of our models with state-of-the-art methods under the self-reconstruction setting. 
TS-Net achieves comparable or better performance when compared with state-of-the-art methods even when using only one subject frame ($K=1$). Though RegionMM \cite{siarohin2021motion} achieves the best performance in the metrics for face videos under the self-reconstruction setting, it performs worse than TS-Net on all the other tasks (\textit{i.e.}, dance videos and cross-identity transfer settings). The reason may be that RegionMM relies on an unsupervised trained region detection network, which may not be robust enough to handle large motions or fine-grained details in various tasks.
In Fig.~\ref{fig:sota}, one can observe that V2V \cite{wang2018video} suffers from color/shape distortion, FS-V2V \cite{wang2019few} misses some details (\textit{e.g.} opened eyes in first row), EDN fails to preserve some details such as in the face or clothing,
FOMM \cite{siarohin2020first} struggles to capture the head/body pose correctly, and RegionMM \cite{siarohin2021motion} generates images with some blurry regions and unrealistic appearance details. (Similar results can also be observed in Fig.~\ref{fig:examples}). In contrast, TS-Net can better handle large motion and preserve identity. Table~\ref{tab:sota} also confirms the effectiveness of using multiple subject frames in TS-Net to collect various appearance information, where most metrics get improved as the number of subject frames increases.
For the cross-identity transfer setting, we conduct a user study to compare models with human perception. As shown in Fig.~\ref{fig:user}, TS-Net gets the most user preference, especially on dance videos. 

\begin{figure}[t]
    \centering
    \includegraphics[width=0.8\linewidth]{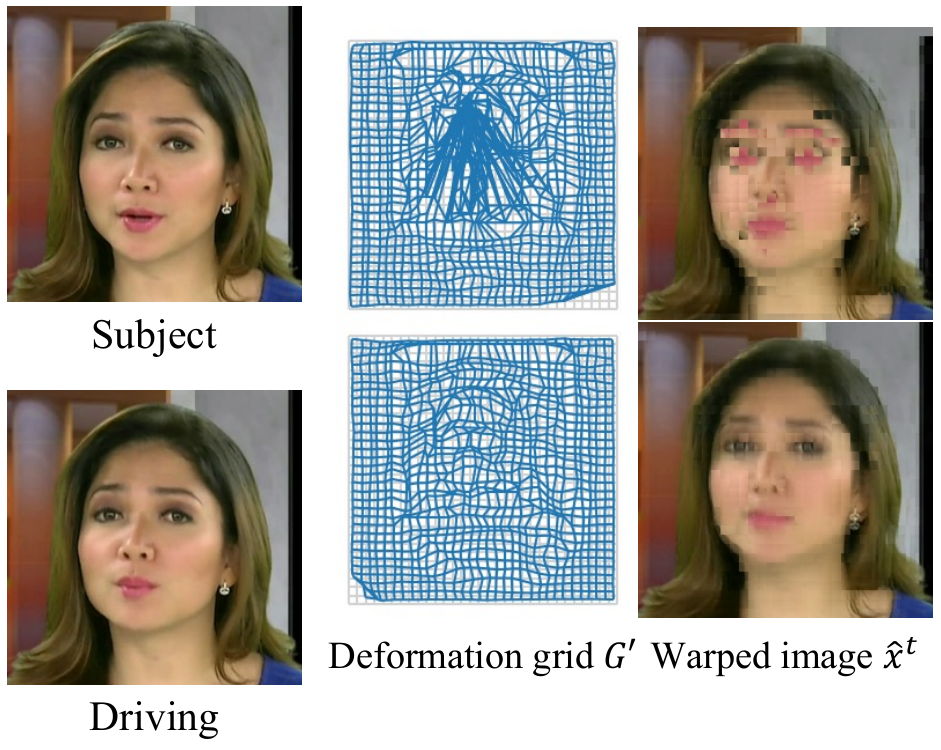}
    \caption{Ablation study for $\mathcal{L}_\text{TRA}$. The top row is from T-Net without using $\mathcal{L}_\text{TRA}$ and the bottom one is with $\mathcal{L}_\text{TRA}$. $\mathcal{L}_\text{TRA}$ clearly learns more reasonable deformation grid and improves warped result.}
    \label{fig:warp_loss}
\vspace{-7pt}
\end{figure}

\begin{figure*}[t]
    \centering
    \vspace{-5pt}
    \includegraphics[width=0.8\textwidth]{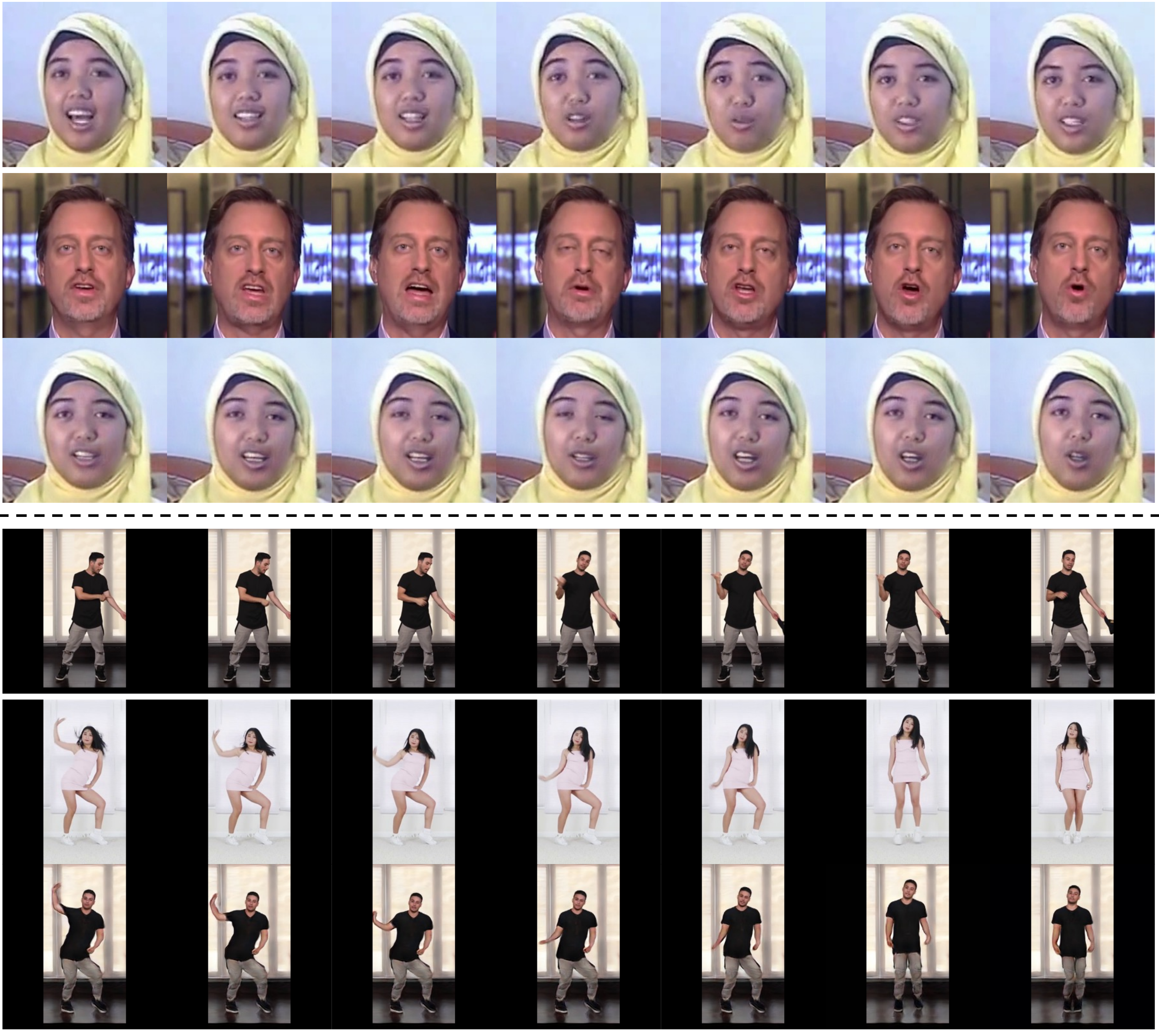}
    \caption{Examples of generated face videos (top block) and dance videos (bottom block) using our proposed TS-Net. For each block, TS-Net synthesizes the new video (3rd row) with the appearance from subject video (1st row) and the motion from driving video (2nd row).}
    \label{fig:video}
    \vspace{-7pt}
\end{figure*}

\noindent
\textbf{Ablation study.} To analyze the effectiveness of each module in TS-Net, we conduct an ablation study on the face video dataset. The number of input subject frames is fixed to be 3 ($K=3$) to ease model training and testing. Table~\ref{tab:aba-face} shows the quantitative comparison results of the ablation study under the self-reconstruction setting. We first train and test two single branch models, T-Net ($\Gamma_\text{tra}$) and S-Net ($\Gamma_\text{syn}$), which only employ the transformation branch or synthesis branch, respectively. From the results shown in Table~\ref{tab:sota} and Table~\ref{tab:aba-face}, one can observe that even a single branch can achieve promising performance when compared to other state-of-the-art methods. However, as shown in Fig.~\ref{fig:aba-face}, warp-based T-Net fails to generate unseen content (\textit{e.g.}, regions marked by red box) while
warp-free S-Net is incompetent to preserve identity. Results demonstrate that a single T-Net or S-Net enables efficient representation learning, and the combination of two branches can complement each other to achieve more satisfactory results.
We also compare the transformation branch trained with and without $\mathcal{L_\text{TRA}}$, T-Net and [T-Net w/o $\mathcal{L_\text{TRA}}$] in Table~\ref{tab:aba-face}, from which one can observe that removing $\mathcal{L_\text{TRA}}$ diminishes performance. As  Fig.~\ref{fig:warp_loss} shows, the lack of $\mathcal{L_\text{TRA}}$ led to a less meaningful deformation grid $G'$ and resulted in a poor warped image $\hat{x}^t$.

\begin{figure}[hb]
    \centering
    \includegraphics[width=\linewidth]{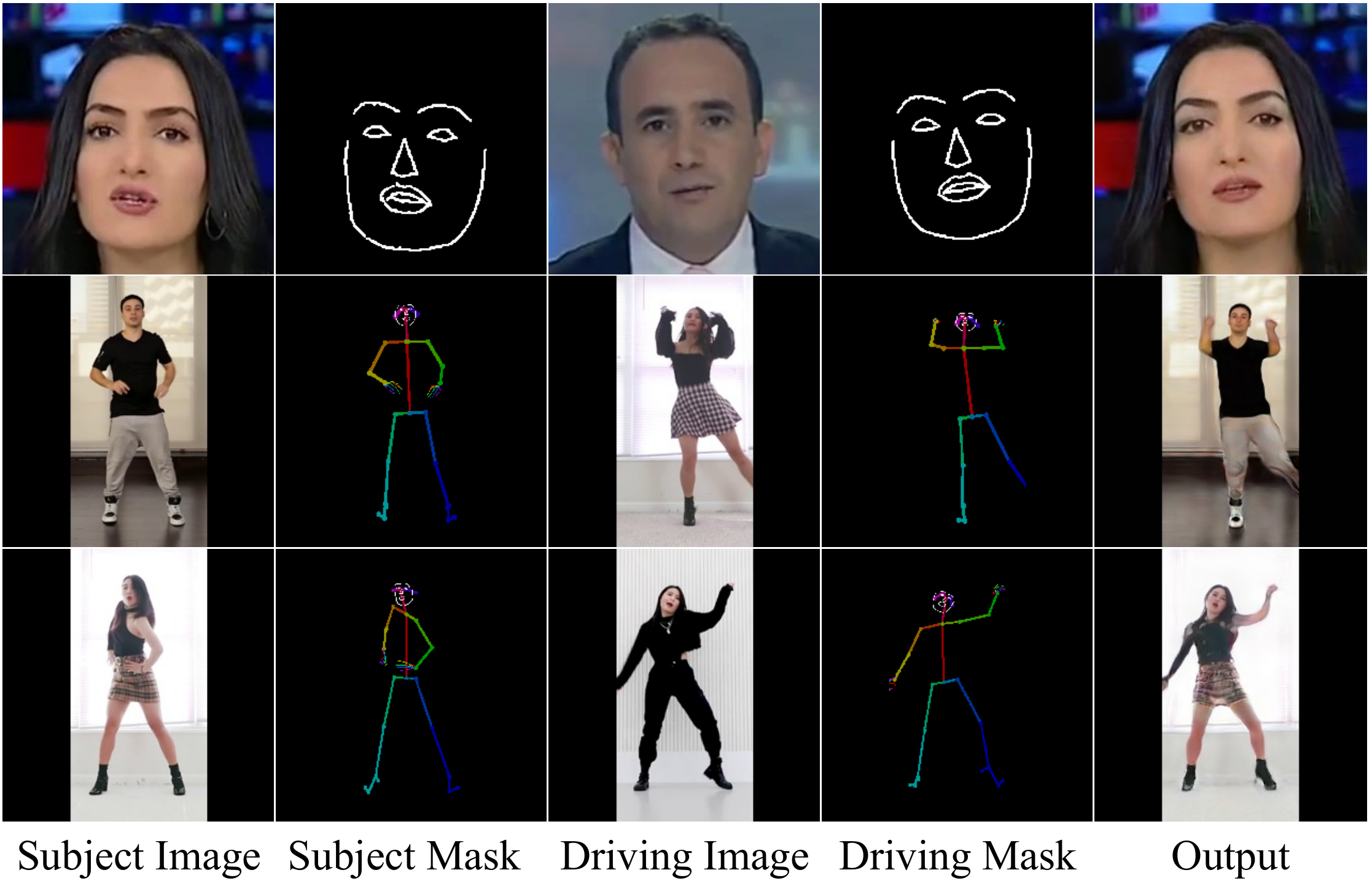}
    \caption{Some failure cases of TS-Net. Driving masks are aligned to match subject masks using the normalization methods in \cite{chan2019everybody,wang2019few}.}
    \label{fig:failure}
\end{figure}

We also evaluate the effectiveness of some common techniques adopted by previous video motion retargeting methods \cite{jeon2020cross,wang2019few,wang2018video}, such as adding cross-identity transfer to the training processing [TS-Net w/ cross] or using the matting function to combine different types of features [TS-Net w/ matting]. To enable cross-identity training, we choose input mask sequence $\mathcal{Z}$ and input image sequence $\mathcal{X}$ from different videos. Thus ground truth frames are not available for training. In this case, we only use adversarial loss $\mathcal{L}_\text{GAN}$ for training, where discriminator $D$ is designed to distinguish the synthesized frame $\hat{x}$ from arbitrary real video frame $x$. For the matting function, we design an extra attention network with similar architecture as fusion network $\Lambda$ to generate a matting mask for combining $\bar{v}^t$ and $\bar{v}^s$. However, both these modules fail to be more effective as Table~\ref{tab:aba-face} shows, which demonstrates that the simple design of TS-Net has already achieved sufficient representation learning and synthesis power.

\begin{table}[t]
\centering
\resizebox{0.7\linewidth}{!}{%
\begin{tabular}{l|c|c}
\hline
Method           & $L_2\downarrow$     & LPIPS $\downarrow$  \\
\hline
T-Net            & 0.0276 & 0.0698 \\
T-Net w/o $\mathcal{L}_\text{TRA}$ & 0.0287 & 0.0725 \\
S-Net            & 0.0285 & 0.0726 \\
\hline
TS-Net w/ cross & 0.0276 & 0.0696 \\
TS-Net w/ matting   & 0.0281 & 0.0696 \\
\hline
TS-Net           & \textbf{0.0271} & \textbf{0.0683} \\
\hline
\end{tabular}%
}
\caption{Ablation Study under the self-reconstruction setting on face dataset. The number of input subject frames is fixed to be 3 ($K=3$).}
\label{tab:aba-face}
\vspace{-5pt}
\end{table}

\section{{Limitations}}
For most cases, our proposed TS-Net can generate realistic videos by only taking 2D sparse masks (see Fig.~\ref{fig:video} and Supp. videos). However, it still suffers from several limitations. First, the input masks of TS-Net are generated from off-the-shelf detectors. Misdetections by the detectors could result in inconsistent motion or incorrect appearances.
As shown in Fig.~\ref{fig:failure}, the synthesized face in the top row has an opened mouth, and the generated man in the middle row shows missing hands. Second, TS-Net sometimes struggles to synthesize high-frequency details.
One can observe a few texture artifacts in the kilt of the last row in Fig.~\ref{fig:failure}. Future work could focus on improving the keypoint detection system and generating more realistic high-frequency textures.

\noindent\textbf{Potential Negative Societal Impact.} Video motion retargeting could be used for unethical purposes \cite{yu2022generating}, \textit{e.g.}, creating videos of celebrities for fake news. We will restrict the usage of our method and model to research purposes only. We also plan to investigate fake video detection techniques \cite{rahman2022qualitative} that will be effective in detecting fake videos like the ones generated by our proposed method. 

\section{{Conclusion}}
In this paper, we propose TS-Net to jointly learn transformation and synthesis for video motion transfer. 
Comprehensive experiments show that TS-Net can achieve state-of-the-art performance on both face and dance videos using only 2D sparse masks. 
In the future, we plan to investigate TS-Net using different kinds of masks and multi-modal information (\textit{e.g.} audio or text) in motion retargeting.

{\small
\bibliographystyle{ieee_fullname}
\bibliography{egbib}

\begin{thebibliography}{10}\itemsep=-1pt

\bibitem{balakrishnan2018synthesizing}
Guha Balakrishnan, Amy Zhao, Adrian~V Dalca, Fredo Durand, and John Guttag.
\newblock Synthesizing images of humans in unseen poses.
\newblock In {\em Proceedings of the IEEE Conference on Computer Vision and
  Pattern Recognition}, pages 8340--8348, 2018.

\bibitem{bansal2018recycle}
Aayush Bansal, Shugao Ma, Deva Ramanan, and Yaser Sheikh.
\newblock Recycle-gan: Unsupervised video retargeting.
\newblock In {\em Proceedings of the European conference on computer vision
  (ECCV)}, pages 119--135, 2018.

\bibitem{cao2019openpose}
Zhe Cao, Gines Hidalgo, Tomas Simon, Shih-En Wei, and Yaser Sheikh.
\newblock Openpose: realtime multi-person 2d pose estimation using part
  affinity fields.
\newblock {\em IEEE transactions on pattern analysis and machine intelligence},
  43(1):172--186, 2019.

\bibitem{chan2019everybody}
Caroline Chan, Shiry Ginosar, Tinghui Zhou, and Alexei~A Efros.
\newblock Everybody dance now.
\newblock In {\em Proceedings of the IEEE/CVF International Conference on
  Computer Vision}, pages 5933--5942, 2019.

\bibitem{chen2020puppeteergan}
Zhuo Chen, Chaoyue Wang, Bo Yuan, and Dacheng Tao.
\newblock Puppeteergan: Arbitrary portrait animation with semantic-aware
  appearance transformation.
\newblock In {\em Proceedings of the IEEE/CVF Conference on Computer Vision and
  Pattern Recognition}, pages 13518--13527, 2020.

\bibitem{chu2020learning}
Mengyu Chu, You Xie, Jonas Mayer, Laura Leal-Taix{\'e}, and Nils Thuerey.
\newblock Learning temporal coherence via self-supervision for gan-based video
  generation.
\newblock {\em ACM Transactions on Graphics (TOG)}, 39(4):75--1, 2020.

\bibitem{deng2009imagenet}
Jia Deng, Wei Dong, Richard Socher, Li-Jia Li, Kai Li, and Li Fei-Fei.
\newblock Imagenet: A large-scale hierarchical image database.
\newblock In {\em 2009 IEEE conference on computer vision and pattern
  recognition}, pages 248--255. Ieee, 2009.

\bibitem{dosovitskiy2015flownet}
Alexey Dosovitskiy, Philipp Fischer, Eddy Ilg, Philip Hausser, Caner Hazirbas,
  Vladimir Golkov, Patrick Van Der~Smagt, Daniel Cremers, and Thomas Brox.
\newblock Flownet: Learning optical flow with convolutional networks.
\newblock In {\em Proceedings of the IEEE international conference on computer
  vision}, pages 2758--2766, 2015.

\bibitem{gafni2021dynamic}
Guy Gafni, Justus Thies, Michael Zollhofer, and Matthias Nie{\ss}ner.
\newblock Dynamic neural radiance fields for monocular 4d facial avatar
  reconstruction.
\newblock In {\em Proceedings of the IEEE/CVF Conference on Computer Vision and
  Pattern Recognition}, pages 8649--8658, 2021.

\bibitem{goodfellow2014generative}
Ian~J Goodfellow, Jean Pouget-Abadie, Mehdi Mirza, Bing Xu, David Warde-Farley,
  Sherjil Ozair, Aaron Courville, and Yoshua Bengio.
\newblock Generative adversarial networks.
\newblock {\em arXiv preprint arXiv:1406.2661}, 2014.

\bibitem{guler2018densepose}
R{\i}za~Alp G{\"u}ler, Natalia Neverova, and Iasonas Kokkinos.
\newblock Densepose: Dense human pose estimation in the wild.
\newblock In {\em Proceedings of the IEEE conference on computer vision and
  pattern recognition}, pages 7297--7306, 2018.

\bibitem{ha2020marionette}
Sungjoo Ha, Martin Kersner, Beomsu Kim, Seokjun Seo, and Dongyoung Kim.
\newblock Marionette: Few-shot face reenactment preserving identity of unseen
  targets.
\newblock In {\em Proceedings of the AAAI Conference on Artificial
  Intelligence}, volume~34, pages 10893--10900, 2020.

\bibitem{he2016deep}
Kaiming He, Xiangyu Zhang, Shaoqing Ren, and Jian Sun.
\newblock Deep residual learning for image recognition.
\newblock In {\em Proceedings of the IEEE conference on computer vision and
  pattern recognition}, pages 770--778, 2016.

\bibitem{hong2022depth}
Fa-Ting Hong, Longhao Zhang, Li Shen, and Dan Xu.
\newblock Depth-aware generative adversarial network for talking head video
  generation.
\newblock In {\em Proceedings of the IEEE/CVF Conference on Computer Vision and
  Pattern Recognition}, pages 3397--3406, 2022.

\bibitem{hu2018videomatch}
Yuan-Ting Hu, Jia-Bin Huang, and Alexander~G Schwing.
\newblock Videomatch: Matching based video object segmentation.
\newblock In {\em Proceedings of the European conference on computer vision
  (ECCV)}, pages 54--70, 2018.

\bibitem{isola2017image}
Phillip Isola, Jun-Yan Zhu, Tinghui Zhou, and Alexei~A Efros.
\newblock Image-to-image translation with conditional adversarial networks.
\newblock In {\em Proceedings of the IEEE conference on computer vision and
  pattern recognition}, pages 1125--1134, 2017.

\bibitem{jaderberg2015spatial}
Max Jaderberg, Karen Simonyan, Andrew Zisserman, et~al.
\newblock Spatial transformer networks.
\newblock {\em Advances in neural information processing systems}, 28, 2015.

\bibitem{jeon2020cross}
Subin Jeon, Seonghyeon Nam, Seoung~Wug Oh, and Seon~Joo Kim.
\newblock Cross-identity motion transfer for arbitrary objects through
  pose-attentive video reassembling.
\newblock In {\em European Conference on Computer Vision}, pages 292--308.
  Springer, 2020.

\bibitem{johnson2016perceptual}
Justin Johnson, Alexandre Alahi, and Li Fei-Fei.
\newblock Perceptual losses for real-time style transfer and super-resolution.
\newblock In {\em European conference on computer vision}, pages 694--711.
  Springer, 2016.

\bibitem{kim2018deep}
Hyeongwoo Kim, Pablo Garrido, Ayush Tewari, Weipeng Xu, Justus Thies, Matthias
  Niessner, Patrick P{\'e}rez, Christian Richardt, Michael Zollh{\"o}fer, and
  Christian Theobalt.
\newblock Deep video portraits.
\newblock {\em ACM Transactions on Graphics (TOG)}, 37(4):1--14, 2018.

\bibitem{kim2019u}
Junho Kim, Minjae Kim, Hyeonwoo Kang, and Kwanghee Lee.
\newblock U-gat-it: unsupervised generative attentional networks with adaptive
  layer-instance normalization for image-to-image translation.
\newblock {\em arXiv preprint arXiv:1907.10830}, 2019.

\bibitem{king2009dlib}
Davis~E King.
\newblock Dlib-ml: A machine learning toolkit.
\newblock {\em The Journal of Machine Learning Research}, 10:1755--1758, 2009.

\bibitem{kingma2014adam}
Diederik~P Kingma and Jimmy Ba.
\newblock Adam: A method for stochastic optimization.
\newblock {\em arXiv preprint arXiv:1412.6980}, 2014.

\bibitem{lin2013network}
Min Lin, Qiang Chen, and Shuicheng Yan.
\newblock Network in network.
\newblock {\em arXiv preprint arXiv:1312.4400}, 2013.

\bibitem{liu2018intriguing}
Rosanne Liu, Joel Lehman, Piero Molino, Felipe~Petroski Such, Eric Frank, Alex
  Sergeev, and Jason Yosinski.
\newblock An intriguing failing of convolutional neural networks and the
  coordconv solution.
\newblock {\em arXiv preprint arXiv:1807.03247}, 2018.

\bibitem{liu2022coordinate}
Yihao Liu, Lianrui Zuo, Shuo Han, Jerry~L Prince, and Aaron Carass.
\newblock Coordinate translator for learning deformable medical image
  registration.
\newblock {\em arXiv preprint arXiv:2203.03626}, 2022.

\bibitem{ma2017pose}
Liqian Ma, Xu Jia, Qianru Sun, Bernt Schiele, Tinne Tuytelaars, and Luc
  Van~Gool.
\newblock Pose guided person image generation.
\newblock {\em arXiv preprint arXiv:1705.09368}, 2017.

\bibitem{mao2017least}
Xudong Mao, Qing Li, Haoran Xie, Raymond~YK Lau, Zhen Wang, and Stephen
  Paul~Smolley.
\newblock Least squares generative adversarial networks.
\newblock In {\em Proceedings of the IEEE international conference on computer
  vision}, pages 2794--2802, 2017.

\bibitem{mathieu2015deep}
Michael Mathieu, Camille Couprie, and Yann LeCun.
\newblock Deep multi-scale video prediction beyond mean square error.
\newblock {\em arXiv preprint arXiv:1511.05440}, 2015.

\bibitem{mirza2014conditional}
Mehdi Mirza and Simon Osindero.
\newblock Conditional generative adversarial nets.
\newblock {\em arXiv preprint arXiv:1411.1784}, 2014.

\bibitem{park2019semantic}
Taesung Park, Ming-Yu Liu, Ting-Chun Wang, and Jun-Yan Zhu.
\newblock Semantic image synthesis with spatially-adaptive normalization.
\newblock In {\em Proceedings of the IEEE/CVF Conference on Computer Vision and
  Pattern Recognition}, pages 2337--2346, 2019.

\bibitem{pumarola2018unsupervised}
Albert Pumarola, Antonio Agudo, Alberto Sanfeliu, and Francesc Moreno-Noguer.
\newblock Unsupervised person image synthesis in arbitrary poses.
\newblock In {\em Proceedings of the IEEE Conference on Computer Vision and
  Pattern Recognition}, pages 8620--8628, 2018.

\bibitem{rahman2022qualitative}
Ashifur Rahman, Md~Mazharul Islam, Mohasina~Jannat Moon, Tahera Tasnim, Nipo
  Siddique, Md Shahiduzzaman, and Samsuddin Ahmed.
\newblock A qualitative survey on deep learning based deep fake video creation
  and detection method.
\newblock {\em Aust. J. Eng. Innov. Technol}, 4(1):13--26, 2022.

\bibitem{ren2021pirenderer}
Yurui Ren, Ge Li, Yuanqi Chen, Thomas~H Li, and Shan Liu.
\newblock Pirenderer: Controllable portrait image generation via semantic
  neural rendering.
\newblock In {\em Proceedings of the IEEE/CVF International Conference on
  Computer Vision}, pages 13759--13768, 2021.

\bibitem{roessler2018faceforensics}
Andreas R\"ossler, Davide Cozzolino, Luisa Verdoliva, Christian Riess, Justus
  Thies, and Matthias Nie{\ss}ner.
\newblock Face{F}orensics: A large-scale video dataset for forgery detection in
  human faces.
\newblock {\em arXiv}, 2018.

\bibitem{sarkar2020neural}
Kripasindhu Sarkar, Dushyant Mehta, Weipeng Xu, Vladislav Golyanik, and
  Christian Theobalt.
\newblock Neural re-rendering of humans from a single image.
\newblock In {\em European Conference on Computer Vision}, pages 596--613.
  Springer, 2020.

\bibitem{siarohin2019appearance}
Aliaksandr Siarohin, St{\'e}phane Lathuili{\`e}re, Enver Sangineto, and Nicu
  Sebe.
\newblock Appearance and pose-conditioned human image generation using
  deformable gans.
\newblock {\em IEEE transactions on pattern analysis and machine intelligence},
  2019.

\bibitem{siarohin2019animating}
Aliaksandr Siarohin, St{\'e}phane Lathuili{\`e}re, Sergey Tulyakov, Elisa
  Ricci, and Nicu Sebe.
\newblock Animating arbitrary objects via deep motion transfer.
\newblock In {\em Proceedings of the IEEE/CVF Conference on Computer Vision and
  Pattern Recognition}, pages 2377--2386, 2019.

\bibitem{siarohin2020first}
Aliaksandr Siarohin, St{\'e}phane Lathuili{\`e}re, Sergey Tulyakov, Elisa
  Ricci, and Nicu Sebe.
\newblock First order motion model for image animation.
\newblock {\em arXiv preprint arXiv:2003.00196}, 2020.

\bibitem{siarohin2018deformable}
Aliaksandr Siarohin, Enver Sangineto, St{\'e}phane Lathuiliere, and Nicu Sebe.
\newblock Deformable gans for pose-based human image generation.
\newblock In {\em Proceedings of the IEEE Conference on Computer Vision and
  Pattern Recognition}, pages 3408--3416, 2018.

\bibitem{siarohin2021motion}
Aliaksandr Siarohin, Oliver~J Woodford, Jian Ren, Menglei Chai, and Sergey
  Tulyakov.
\newblock Motion representations for articulated animation.
\newblock In {\em Proceedings of the IEEE/CVF Conference on Computer Vision and
  Pattern Recognition}, pages 13653--13662, 2021.

\bibitem{simonyan2014very}
Karen Simonyan and Andrew Zisserman.
\newblock Very deep convolutional networks for large-scale image recognition.
\newblock {\em arXiv preprint arXiv:1409.1556}, 2014.

\bibitem{song2019unsupervised}
Sijie Song, Wei Zhang, Jiaying Liu, and Tao Mei.
\newblock Unsupervised person image generation with semantic parsing
  transformation.
\newblock In {\em Proceedings of the IEEE/CVF Conference on Computer Vision and
  Pattern Recognition}, pages 2357--2366, 2019.

\bibitem{tao2022structure}
Jiale Tao, Biao Wang, Borun Xu, Tiezheng Ge, Yuning Jiang, Wen Li, and Lixin
  Duan.
\newblock Structure-aware motion transfer with deformable anchor model.
\newblock In {\em Proceedings of the IEEE/CVF Conference on Computer Vision and
  Pattern Recognition}, pages 3637--3646, 2022.

\bibitem{ulyanov2016instance}
Dmitry Ulyanov, Andrea Vedaldi, and Victor Lempitsky.
\newblock Instance normalization: The missing ingredient for fast stylization.
\newblock {\em arXiv preprint arXiv:1607.08022}, 2016.

\bibitem{wang2019few}
Ting-Chun Wang, Ming-Yu Liu, Andrew Tao, Guilin Liu, Jan Kautz, and Bryan
  Catanzaro.
\newblock Few-shot video-to-video synthesis.
\newblock {\em arXiv preprint arXiv:1910.12713}, 2019.

\bibitem{wang2018video}
Ting-Chun Wang, Ming-Yu Liu, Jun-Yan Zhu, Guilin Liu, Andrew Tao, Jan Kautz,
  and Bryan Catanzaro.
\newblock Video-to-video synthesis.
\newblock {\em arXiv preprint arXiv:1808.06601}, 2018.

\bibitem{wang2018high}
Ting-Chun Wang, Ming-Yu Liu, Jun-Yan Zhu, Andrew Tao, Jan Kautz, and Bryan
  Catanzaro.
\newblock High-resolution image synthesis and semantic manipulation with
  conditional gans.
\newblock In {\em Proceedings of the IEEE conference on computer vision and
  pattern recognition}, pages 8798--8807, 2018.

\bibitem{wang2021one}
Ting-Chun Wang, Arun Mallya, and Ming-Yu Liu.
\newblock One-shot free-view neural talking-head synthesis for video
  conferencing.
\newblock In {\em Proceedings of the IEEE/CVF Conference on Computer Vision and
  Pattern Recognition}, pages 10039--10049, 2021.

\bibitem{wiles2018x2face}
Olivia Wiles, A Koepke, and Andrew Zisserman.
\newblock X2face: A network for controlling face generation using images,
  audio, and pose codes.
\newblock In {\em Proceedings of the European conference on computer vision
  (ECCV)}, pages 670--686, 2018.

\bibitem{xu2013survey}
Chang Xu, Dacheng Tao, and Chao Xu.
\newblock A survey on multi-view learning.
\newblock {\em arXiv preprint arXiv:1304.5634}, 2013.

\bibitem{yang2020transmomo}
Zhuoqian Yang, Wentao Zhu, Wayne Wu, Chen Qian, Qiang Zhou, Bolei Zhou, and
  Chen~Change Loy.
\newblock Transmomo: Invariance-driven unsupervised video motion retargeting.
\newblock In {\em Proceedings of the IEEE/CVF Conference on Computer Vision and
  Pattern Recognition}, pages 5306--5315, 2020.

\bibitem{yu2022generating}
Sihyun Yu, Jihoon Tack, Sangwoo Mo, Hyunsu Kim, Junho Kim, Jung-Woo Ha, and
  Jinwoo Shin.
\newblock Generating videos with dynamics-aware implicit generative adversarial
  networks.
\newblock {\em arXiv preprint arXiv:2202.10571}, 2022.

\bibitem{zakharov2019few}
Egor Zakharov, Aliaksandra Shysheya, Egor Burkov, and Victor Lempitsky.
\newblock Few-shot adversarial learning of realistic neural talking head
  models.
\newblock In {\em Proceedings of the IEEE/CVF International Conference on
  Computer Vision}, pages 9459--9468, 2019.

\bibitem{zhang2020cross}
Pan Zhang, Bo Zhang, Dong Chen, Lu Yuan, and Fang Wen.
\newblock Cross-domain correspondence learning for exemplar-based image
  translation.
\newblock In {\em Proceedings of the IEEE/CVF Conference on Computer Vision and
  Pattern Recognition}, pages 5143--5153, 2020.

\bibitem{zhang2018perceptual}
Richard Zhang, Phillip Isola, Alexei~A Efros, Eli Shechtman, and Oliver Wang.
\newblock The unreasonable effectiveness of deep features as a perceptual
  metric.
\newblock In {\em CVPR}, 2018.

\bibitem{zhu2017unpaired}
Jun-Yan Zhu, Taesung Park, Phillip Isola, and Alexei~A Efros.
\newblock Unpaired image-to-image translation using cycle-consistent
  adversarial networks.
\newblock In {\em Proceedings of the IEEE international conference on computer
  vision}, pages 2223--2232, 2017.

\end{thebibliography}
}

\end{document}